\pdfmapfile{+cvpr-font.map}

\documentclass[10pt,twocolumn,letterpaper]{article}

\usepackage[pagenumbers]{cvpr}

\usepackage{microtype}

\makeatletter
\renewcommand\paragraph{\@startsection{paragraph}{4}{\z@}
  {.5em}{-1em}{\normalfont\normalsize\bfseries}}
\makeatother

\usepackage{mathtools}
\usepackage{multirow}
\usepackage{array}
\usepackage{tabularx}
\usepackage{tgheros}
\usepackage{algorithm}
\usepackage{algpseudocode}

\newcommand{\method}{\textup{\textsc{GaussLearner}}\xspace}

\newcolumntype{L}[1]{>{\raggedright\arraybackslash}p{#1}}
\newcolumntype{C}[1]{>{\centering\arraybackslash}p{#1}}

\definecolor{cvprblue}{rgb}{0.21,0.49,0.74}
\usepackage[breaklinks,colorlinks,allcolors=cvprblue]{hyperref}

\def\confName{CVPR}
\def\confYear{2027}

\title{GaussLearner: 3D Gaussian Primitives as Self-Regulating Continual Learners}

\author{%
Gaoxiang Linghu$^{1}$\quad
Haonan Zhou$^{1}$\quad
Xuanhe Liu$^{1}$\quad
Weili Xu$^{1}$\\[0.4em]
Kaiyue Zhou$^{2}$$^{*}$\quad
Jun Xiao$^{1}$\quad
Gaoang Wang$^{1}$$^{*}$\\[0.6em]
{\normalsize $^{1}$Zhejiang University\quad
$^{2}$Chengdu Minto Tech}\\[0.2em]
{\footnotesize $^{*}$Corresponding authors.}%
}
\hypersetup{%
  pdftitle={GaussLearner: 3D Gaussian Primitives as Self-Regulating Continual Learners},
  pdfauthor={Gaoxiang Linghu, Haonan Zhou, Xuanhe Liu, Weili Xu, Kaiyue Zhou, Jun Xiao, Gaoang Wang},
  pdfsubject={Continual 3D Gaussian reconstruction}
}

\makeatletter
\patchcmd{\@maketitle}{\newpage}{\newpage\vspace*{-25.298pt}}{}{}
\patchcmd{\@maketitle}{\vskip .375in}{\vskip -13pt}{}{}
\patchcmd{\@maketitle}{\vspace*{24pt}}{\vspace*{10pt}}{}{}
\patchcmd{\endabstract}{\vspace*{12pt}}{\vspace*{4pt}}{}{}
\apptocmd{\@maketitle}{%
    \vspace*{-0.5em}
    \begin{center}
        \begin{minipage}{\textwidth}
            \centering
            \includegraphics[width=\textwidth]{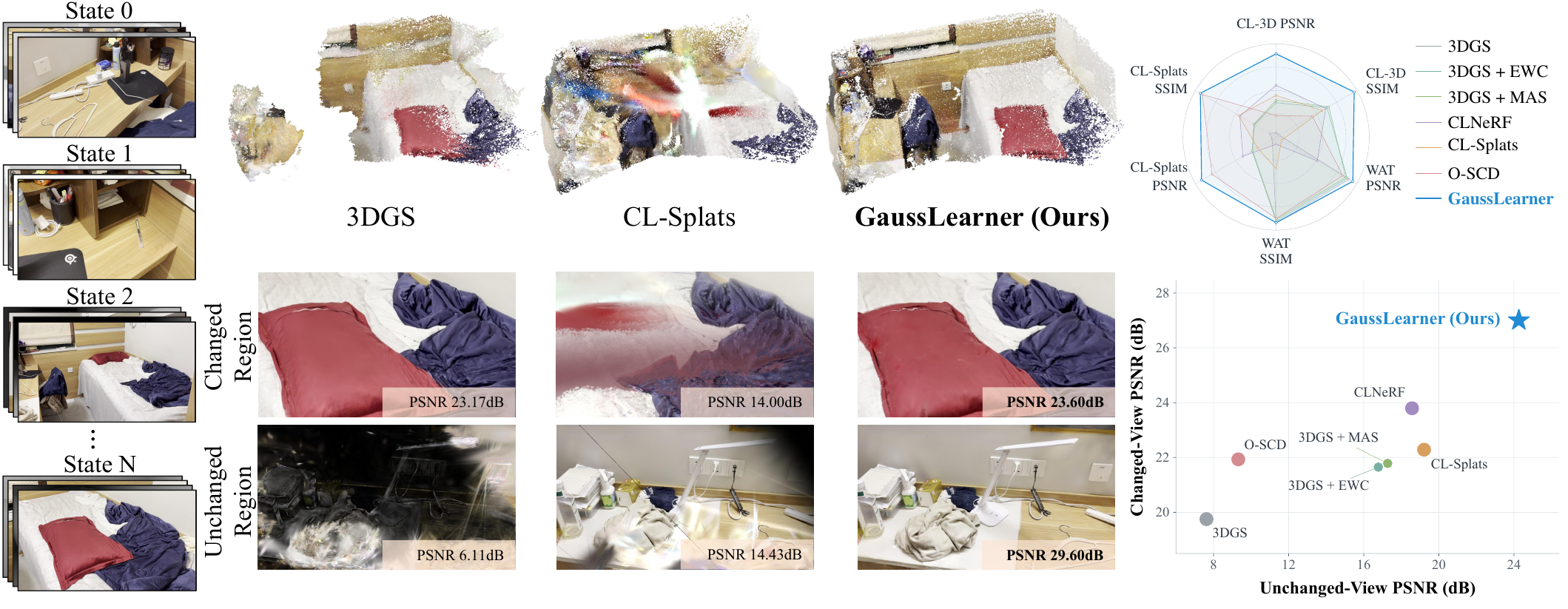}
            \captionsetup{hypcap=false}
            \captionof{figure}{\textbf{Continual 3D Gaussian reconstruction across evolving scene states.}
            Directly fine-tuning an existing Gaussian field~\cite{kerbl2023gaussians} leads to catastrophic forgetting of previously reconstructed regions, while selective local
            optimization methods~\cite{ackermann2025clsplats} struggle to balance adaptation with preservation.\ \method{} reconstructs
            both changed and unchanged regions and improves reconstruction quality across datasets.}\label{fig:teaser}
        \end{minipage}
    \end{center}
    \vspace*{0.5em}
}{}{}
\AddToHook{cmd/@makecol/before}{%
  \ifnum\c@page=8
    \if@firstcolumn\raggedbottom\else\flushbottom\fi
  \else
    \raggedbottom
  \fi
}
\makeatother

\begin{document}
\maketitle
\raggedbottom%
\begin{abstract}
Continual learning of 3D Gaussian fields aims to maintain faithful reconstructions of the evolving world.
However, directly fine-tuning an existing Gaussian field on new observations can cause catastrophic forgetting of previously reconstructed content.
To address this challenge, existing methods employ selective local optimization, replay, or parameter regularization, yet still struggle to balance adaptation to scene changes with preservation of still-valid content.
Our key insight is to view 3D Gaussian primitives themselves as self-regulating continual learners.
Based on this insight, we introduce \method, which equips each Gaussian with a closed continual-learning loop connecting observation, self-regulation, and adaptation at the primitive level.
At each novel state, \method first attributes pixel-wise inconsistencies in current observations to their contributing Gaussians.
Each Gaussian then self-regulates through a learning weight that balances current observations against cross-state reliability.
This weight governs both Gaussian parameter adaptation and optimizer memory refresh.
As current inconsistencies are resolved, the adaptation outcome renews each Gaussian's reliability for future states.
The resulting field is a globally coupled population of locally self-regulating learners.
Across scenes from CL-Splats, WAT, and our CL-3D dataset, \method improves average PSNR by 3.41 dB over the strongest baseline. On CL-3D, it achieves the best reconstruction quality on both changed and unchanged views.
\end{abstract}

\section{Introduction}
\label{sec:intro}

Continual learning for 3D scenes aims to update a scene representation as new observations keep arriving~\cite{li2016learning,schmid2024khronos}.
These observations can reveal newly explored regions~\cite{yang2025cgs}, rearranged objects~\cite{wu2026scenediff}, or changes in scene appearance~\cite{cai2023clnerf}.
The central challenge is to balance adaptation to scene changes with preservation of still-valid scene content~\cite{schmid2022panoptic}.
This capability is essential for long-term robotic navigation~\cite{chhablani2025embodiedsplat}, AR/VR~\cite{sun2025lts,vachha2026splatxr}, and embodied intelligence~\cite{zhang2026glmap}.

Existing methods address this challenge through selective local optimization, replay, or parameter regularization.
CL-Splats and O-SCD first detect changed regions in 2D observations, lift them into 3D, and restrict optimization to the selected Gaussians~\cite{ackermann2025clsplats,galappaththige2026oscd}.
They therefore depend heavily on 2D change detection~\cite{galappaththige2025mv3dcd} and become fragile under illumination variation or occlusion~\cite{sachdeva2023change3d}.
GaussianUpdate instead employs visibility-aware generative replay~\cite{zeng2025gaussianupdate}.
More broadly, replay- and regularization-based approaches preserve historical knowledge through additional supervision or constraints~\cite{cai2023clnerf,wu2023clnerf,yang2025cgs}, but they can retain obsolete content or resist necessary changes.

\begin{figure}[t]
    \centering
    \includegraphics[width=\columnwidth]{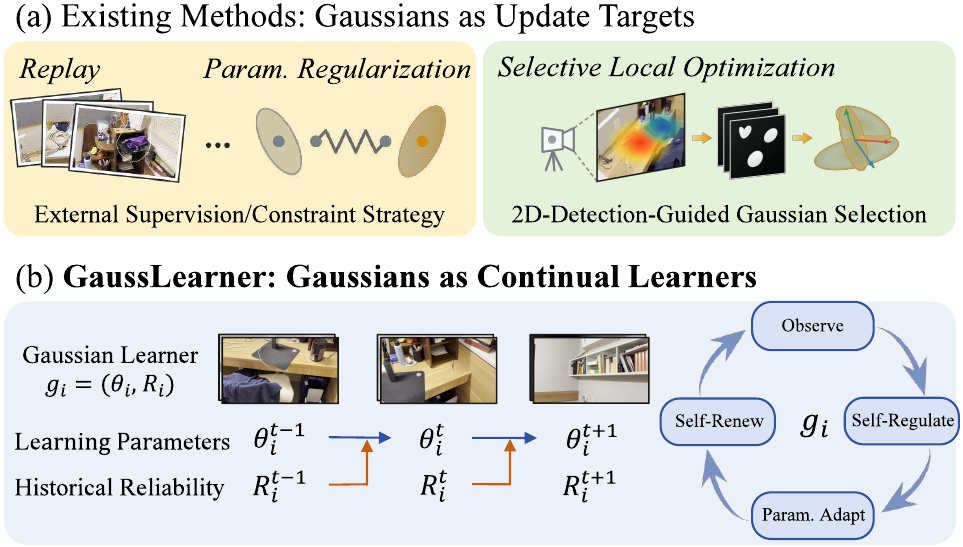}
    \caption{\textbf{Comparison of 3D Gaussian continual learning paradigms.}
    (a) Existing methods treat Gaussians as update targets.
    Replay and regularization may retain obsolete content or hinder adaptation, while selective local optimization can propagate 2D detection errors into 3D updates.
    (b) \method instead views each Gaussian primitive as a continual learner equipped with a closed learning loop.}
    \label{fig:gaussian-learners}
\end{figure}

Our central insight is to view 3D Gaussian primitives as self-regulating continual learners.
In prior paradigms, a continual-learning policy acts on the Gaussian field, with individual Gaussians serving as the update targets.
As shown in Fig.~\ref{fig:gaussian-learners}, we instead aim to equip each Gaussian with a closed continual-learning loop,
in which each primitive receives attributed observations, self-regulates through its learning weight, and adapts through coordinated parameter updates and optimizer-memory refresh.

Based on this insight, we propose \method, a continual-learning framework that enables every Gaussian to self-regulate its adaptation across scene states.
At each scene state, \method first attributes pixel-wise inconsistencies in current observations to the contributing Gaussians.
Each Gaussian learner then weighs its attributed observations against cross-state reliability to form a per-Gaussian learning weight.
This weight governs both how far the Gaussian follows its current Adam proposal and how strongly its optimizer memory is refreshed.
As current inconsistencies are resolved, the adaptation outcome renews the Gaussian's cross-state reliability for future scene states.
The resulting field is a globally coupled population of locally self-regulating continual learners.

We evaluate \method on CL-Splats, WAT, and our self-captured CL-3D dataset.
Across all scenes, \method improves PSNR by 3.41 dB on average over the strongest baseline.
It also achieves the best reconstruction quality on both changed and unchanged views on the CL-3D dataset, demonstrating its ability to adapt to revised scene content while preserving content that remains valid.

Overall, our key contributions are:
\begin{itemize}
    \item We propose \method, a continual-learning framework that views 3D Gaussian primitives as continual learners, thereby enabling effective scene adaptation while mitigating catastrophic forgetting.
    \item We develop a self-regulating Gaussian optimization process in which a per-Gaussian learning weight jointly governs parameter adaptation and optimizer-memory refresh.
    \item We conduct extensive experiments across datasets, demonstrating that \method effectively adapts to changed content while preserving still-valid content.
\end{itemize}

\section{Related Work}
\label{sec:related}

\subsection{Continual Learning of 3D Gaussian Scenes}

Continual learning of 3D Gaussian scenes updates a reconstructed field across discrete scene states while preserving still-valid content~\cite{cross2026temporal}.
This differs from online Gaussian SLAM, which integrates streaming observations into a current map~\cite{keetha2024splatam,matsuki2024gaussianslam,yan2024gsslam,huang2024photoslam,yugay2026game,peng2024rtgslam,zhu2025loopsplat}.
Multi-state methods mainly use selective local optimization~\cite{zhou2026scar3d,lu2025gschange}, replay, or parameter regularization.
CL-Splats~\cite{ackermann2025clsplats} and O-SCD~\cite{galappaththige2026oscd} detect changes in 2D before updating selected Gaussians.
GaussianUpdate~\cite{zeng2025gaussianupdate} uses generative replay, which reduces forgetting but adds overhead and may retain obsolete content.
CGS~\cite{yang2025cgs} protects historically important Gaussians through regularization, which can resist necessary changes.
Other work explores probabilistic updates and scene memories~\cite{vandemaele2024vbgs,xu2026selfevolving,kim2026ltgs,mazur2026primitive}.
Appendix~F provides a unified view of field-level continual Gaussian Splatting methods, highlighting our shift to per-Gaussian self-regulation.
\method coordinates parameter adaptation and optimizer-memory refresh through a learning weight determined by current observations and cross-state reliability.

\subsection{Continual Learning for Implicit Mapping}

Implicit maps encode geometry and appearance in shared parameters, so fitting new observations can disturb unobserved regions~\cite{mildenhall2020nerf,azinovic2022neuralrgbd,sucar2021imap,zhu2022niceslam,wang2023coslam,yang2022voxfusion}.
Existing methods use replay~\cite{yan2021cnm,cai2023clnerf,ortiz2022isdf}, parameter regularization or isolation~\cite{kirkpatrick2017ewc,aljundi2018mas,mallya2018packnet,wu2023clnerf}, and distillation~\cite{guo2024unikd}.
These strategies reduce forgetting but may retain outdated scene content.
Distributed mapping reconciles models without sharing raw data~\cite{asadi2024dinerf,deng2024macim}, while TACO~\cite{zhou2026taco} uses temporal consensus to preserve reliable history and revise conflicting regions.
\method pursues the same adaptation and preservation balance in an explicit field, but makes each Gaussian a stateful learner that regulates its own adaptation.

\begin{figure*}[t]
    \centering
    \includegraphics[width=0.95\textwidth]{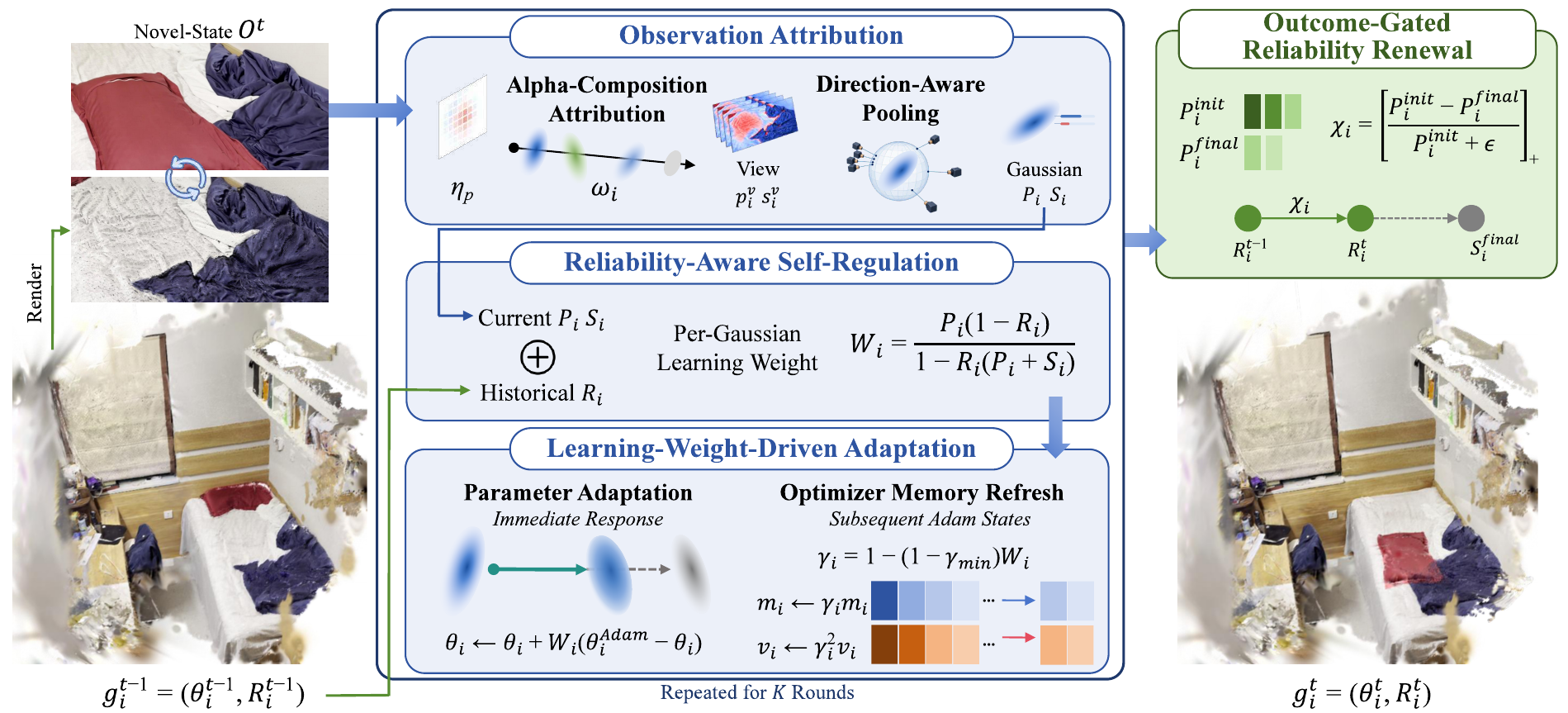}
    \caption{\textbf{Framework of \method.}
    Each Gaussian acts as a stateful continual learner with scene parameters and cross-state reliability.
    At each scene state, pixel-wise inconsistencies in current observations are attributed to their contributing Gaussians.
    Each Gaussian then self-regulates through a per-Gaussian learning weight that balances current observations against cross-state reliability.
    This weight governs both parameter adaptation and optimizer-memory refresh.
    Cross-state reliability is updated at state boundaries using resolved pressure and final support.
    The resulting field is a globally coupled population of locally self-regulating continual learners.}
    \label{fig:framework}
\end{figure*}

\section{Method}
\label{sec:method}

\subsection{3D Gaussian Primitives as Continual Learners}
\label{sec:overview}

Following~\cite{ackermann2025clsplats,zeng2025gaussianupdate,yang2025cgs}, we consider
continual learning of 3D Gaussian fields over a sequence of discrete scene
states $t=0,\ldots,T$.
At each subsequent state $t\geq 1$, the system receives registered RGB-D observations $\mathcal O^t$.
The objective is to update the previously reconstructed field $\mathcal G^{t-1}$ into
$\mathcal G^t$, incorporating changed content while
preserving previously reconstructed content that remains valid.

As illustrated in Fig.~\ref{fig:framework}, \method makes the Gaussian
primitives themselves continual learners. We
represent each Gaussian learner as $g_i=(\boldsymbol\theta_i,R_i)$, where
$\boldsymbol\theta_i$ contains its standard 3DGS parameters and
$R_i\in[0,1)$ denotes the \textbf{cross-state reliability}.
We introduce $R_i$ to carry the Gaussian's established support across scene states.
At scene state $t$, each learner undergoes the transition
\begin{equation}
    (\boldsymbol\theta_i^t,R_i^t)
    =
    \mathcal U\!\left((\boldsymbol\theta_i^{t-1},R_i^{t-1}),\mathcal O^t\right),
    \label{eq:gaussian-transition}
\end{equation}
where $\mathcal U$ denotes the complete update process of \method.
Within each scene state, cross-state reliability remains fixed while the
Gaussian learners adapt over $K$ rounds. At the beginning
of every round, current observations are attributed to their contributing
Gaussians, as detailed in Section~\ref{sec:attribution}.
We weigh attributed pressure against cross-state reliability to form a learning weight for each Gaussian learner.
Section~\ref{sec:self-regulation} details how this weight is formed, while
Section~\ref{sec:adaptation} describes how it governs the Gaussian's parameter
response and optimizer-memory refresh.
After the final round, the adaptation outcome renews its cross-state
reliability for future learning in Section~\ref{sec:closed-loop}.

\subsection{Pixel-to-Gaussian Observation Attribution}
\label{sec:attribution}
Each Gaussian learner begins self-regulation by observing whether a new scene
state supports its current representation or pressures it to change. Given RGB-D
observations of the new state, we first attribute pixel-wise rendering inconsistencies
to the contributing Gaussians and then pool the resulting support and pressure across views.

\paragraph{Alpha-Composition Attribution.}
For each incoming viewpoint, we render RGB and depth from the current Gaussian
field and compare them with the observation to define a bounded rendering
inconsistency:
\begin{equation}
    \eta_p
    =1-\exp\!\left(-\frac{\Delta_p^{\mathrm{RGB}}
                         +\lambda_D\Delta_p^D}{1+\lambda_D}\right).
    \label{eq:rendering-inconsistency}
\end{equation}
Here, $\Delta_p^{\mathrm{RGB}}$ and $\Delta_p^D$ denote calibrated RGB and
log-depth differences, respectively.

Starting from similar pixel-wise photometric or feature differences~\cite{oquab2024dinov2,xiong2024efficientsam,ravi2025sam2}, methods based on selective local optimization
transform them into binary change masks to select Gaussians for subsequent local optimization~\cite{ackermann2025clsplats,galappaththige2026oscd,kim2026ltgs,zhou2026scar3d}.
We argue that this paradigm reduces continuous observations to hard decisions and directly
propagates errors in 2D change detection into the Gaussian field.

\method instead attributes $\eta_p$ according to the actual rendering contribution of each
Gaussian, which is exposed by the standard Gaussian Splatting
renderer~\cite{kerbl2023gaussians} through its alpha-composition weights.
For Gaussian $i$, its rendering contribution at pixel $p$ is
\begin{equation}
    \omega_{ip}=\alpha_{ip}\prod_{j<i}(1-\alpha_{jp}),
    \label{eq:alpha-attribution}
\end{equation}
Here, $\alpha_{ip}$ is its projected opacity, and $j<i$ follows the rendering
depth order.
We aggregate the inconsistencies into view-wise support and pressure:
\begin{equation}
    s_i^v=\sum_{p\in v}\omega_{ip}(1-\eta_p),
    \qquad
    p_i^v=\sum_{p\in v}\omega_{ip}\eta_p.
    \label{eq:pixel-attribution}
\end{equation}
Support $s_i^v$ and pressure $p_i^v$ accumulate Gaussian $i$'s contributions to
consistent and inconsistent pixels, respectively. This renderer-native
attribution routes pixel-wise observations through the renderer's own
visibility and occlusion weights. Specifically, we differentiate a weighted
feature-rendering readout~\cite{joseph2025gradient} with respect to two-channel Gaussian probes
$\mathbf f_i$ to obtain per-Gaussian support and pressure, avoiding an explicit Gaussian-by-pixel weight tensor.

\paragraph{Direction-Aware Pooling.}
Support and pressure from a single view reflect only one observation direction.
Nearby viewpoints often provide redundant visibility, whereas distinct
directions offer complementary coverage.
Inspired by~\cite{xue2026gavis}, we introduce a Direction-Aware
Pooling operator to organize observations across views.
\begin{equation}
    (S_i,P_i)
    =\operatorname{DirPool}\!\left(\{(s_i^v,p_i^v,\mathbf d_i^v)\}_v\right).
    \label{eq:direction-pooling}
\end{equation}
Here, $\mathbf d_i^v=(\mathbf c_v-\boldsymbol\mu_i)/
\|\mathbf c_v-\boldsymbol\mu_i\|_2$ is the viewing direction from Gaussian
center $\boldsymbol\mu_i$ to camera center $\mathbf c_v$.
Specifically, $\operatorname{DirPool}(\cdot)$ 
softly assigns each view to $G$ approximately uniform Fibonacci anchors using a von Mises--Fisher angular
kernel~\cite{banerjee2005vmf}. It computes weighted averages within each anchor,
aggregates them according to directional coverage, and jointly normalizes the
result into per-Gaussian support $S_i$ and pressure $P_i$. The resulting quantities satisfy
$S_i\geq 0$, $P_i\geq 0$, while $S_i+P_i<1$ measures the
current observation coverage of Gaussian $i$. 
After the initial reconstruction, we set $R_i^0=S_i^0$ to initialize cross-state reliability.
In subsequent scene states, $S_i$ and $P_i$ are recomputed at the beginning of
every adaptation round to remain aligned with the evolving Gaussian field.

\subsection{Reliability-Aware Self-Regulation}
\label{sec:self-regulation}

We determine how strongly each Gaussian learner should respond to current inconsistencies, considering both the available observations and its learning history.
We balance current observations against cross-state reliability
through an observation weight $\pi_i^{\mathrm{obs}}$ and a historical weight
$\pi_i^{\mathrm{hist}}$:
\begin{equation*}
    \pi_i^{\mathrm{obs}}=\frac{1}{1-S_i-P_i},
    \qquad
    \pi_i^{\mathrm{hist}}=\frac{R_i}{1-R_i}.
\end{equation*}
The observation weight $\pi_i^{\mathrm{obs}}$ increases with current coverage, while the historical
weight $\pi_i^{\mathrm{hist}}$ is the odds of reliability. Taking the same normalized form as the scalar
Kalman gain expressed in precisions~\cite{kalman1960new}, the adaptation share
$\beta_i$ scales attributed pressure into the learning weight:
\begin{align}
        \beta_i
        &=\frac{\pi_i^{\mathrm{obs}}}
        {\pi_i^{\mathrm{obs}}+\pi_i^{\mathrm{hist}}},\notag\\
        W_i
        &=P_i\beta_i
        =\frac{P_i(1-R_i)}{1-R_i(S_i+P_i)}.
    \label{eq:learning-weight}
\end{align}
Further analysis is provided in Appendix~A.4.
Pressure $P_i$ thus provides the drive to change, while $\beta_i$ determines how
strongly current observations influence adaptation relative to historical
reliability. Without established history ($R_i=0$), $\beta_i=1$ and $W_i=P_i$.
At fixed reliability, broader coverage increases $\beta_i$; at fixed coverage,
higher reliability reduces it. An unobserved Gaussian receives no update because
$P_i=0$. Support therefore modulates the response to pressure but never triggers
adaptation on its own.

\begin{figure*}[t]
    \centering
    \includegraphics[width=\textwidth]{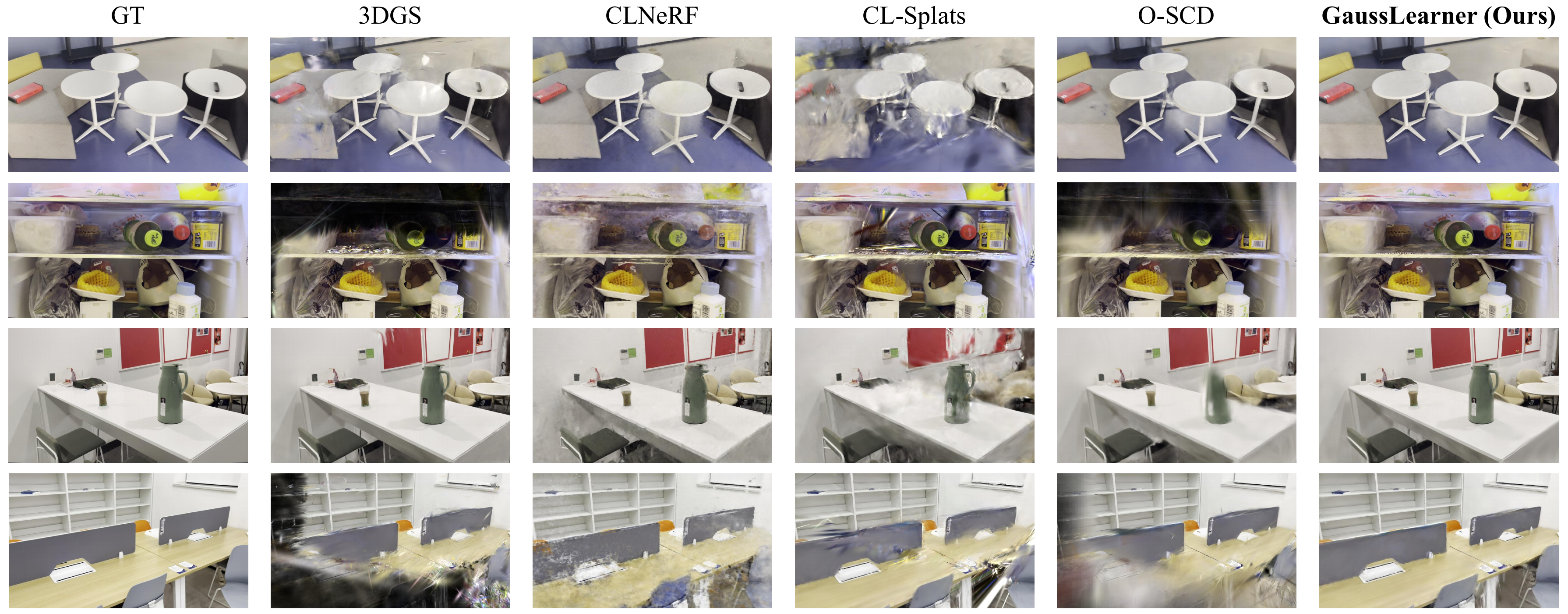}
    \caption{\textbf{Qualitative comparisons on CL-3D.}
    From top to bottom are ClubRoom, Fridge, OfficeHub, and StudyNook.
    \method adapts to revised content while preserving still-valid regions,
    with less forgetting and fewer reconstruction artifacts.}
    \label{fig:rivalsplats-qualitative}
\end{figure*}

\subsection{Learning-Weight-Driven Gaussian Adaptation}
\label{sec:adaptation}

The resulting learning weight $W_i$ translates each Gaussian's self-regulation
into two complementary adaptation mechanisms that control its immediate
parameter response and refresh optimizer memory for subsequent rounds.

\paragraph{Weight-Scaled Parameter Adaptation.}
At the parameter level, self-regulation concerns how much of an optimizer update
is actually realized.
Scaling a gradient does not generally scale an Adam update by the same factor,
because Adam normalizes it using first- and second-order
moments~\cite{kingma2015adam}.
Thus, we separate proposal construction from proposal acceptance: Adam first
constructs a complete parameter proposal under the reconstruction objective
$\mathcal L$, and $W_i$ determines how much of that proposal is realized:
\begin{align}
    \widetilde{\boldsymbol\theta}_i
    &=\operatorname{Adam}
    (\boldsymbol\theta_i,
      \nabla_{\boldsymbol\theta_i}\mathcal L),\nonumber\\
    \boldsymbol\theta_i
    &\leftarrow\boldsymbol\theta_i
    +W_i(\widetilde{\boldsymbol\theta}_i-
    \boldsymbol\theta_i).
    \label{eq:proposal-acceptance}
\end{align}
A small $W_i$ realizes only a limited portion of the proposal, whereas a value
near one accepts it almost completely.
The same $W_i$ modulates position, scale, rotation, opacity, and appearance,
with normalized interpolation applied to rotations.
Each Gaussian is therefore updated coherently as a complete 3D primitive.

\paragraph{Weight-Guided Optimizer Memory Refresh.}
Self-regulation also extends beyond the current parameter adaptation to the
optimizer history carried into subsequent rounds.
Larger learning weights call for stronger adaptation.
We therefore attenuate accumulated Adam moments more strongly for these Gaussians, giving subsequent gradients greater influence.
Smaller weights retain more optimizer history as the need for adaptation decreases.
Accordingly, $W_i$ defines a memory-retention factor $\gamma_i$ that governs the refresh of Adam's first-moment estimate $\mathbf m_i$ and second-moment estimate $\mathbf v_i$:
\begin{equation}
    \gamma_i=1-(1-\gamma_{\min})W_i,
    \quad
    \mathbf m_i\leftarrow\gamma_i\mathbf m_i,
    \quad
    \mathbf v_i\leftarrow\gamma_i^2\mathbf v_i,
    \label{eq:memory-refresh}
\end{equation}
Both moment estimates are maintained for all parameter fields in
$\boldsymbol\theta_i$, and $\gamma_{\min}$ is the minimum retention factor.
A large $W_i$ produces a smaller $\gamma_i$ and a stronger memory refresh.
The paired scaling of $\mathbf m_i$ and $\mathbf v_i$ approximately preserves
the instantaneous normalized Adam direction, since
$\gamma_i\mathbf m_i/\sqrt{\gamma_i^2\mathbf v_i}
=\mathbf m_i/\sqrt{\mathbf v_i}$ when the numerical stabilizer is omitted~\cite{kingma2015adam}.
Recomputing $W_i$ each round allows the refresh to respond to
evolving pressure and weaken as current inconsistencies are resolved.

Together, the same $W_i$ carries Gaussian self-regulation across two
complementary horizons: it controls how far a Gaussian follows its current Adam
proposal and how strongly its optimizer history shapes subsequent updates.

\subsection{Outcome-Gated Reliability Renewal}
\label{sec:closed-loop}

Within each state, reliability remains fixed. At state boundaries, adaptation
outcomes renew $\pi_i^{\mathrm{hist}}$, gated by the resolved fraction $\chi_i$
of initial pressure:
\begin{equation}
    \begin{aligned}
        \chi_i
        &=\left[\frac{P_i^{\mathrm{init}}-P_i^{\mathrm{final}}}
        {P_i^{\mathrm{init}}+\epsilon}\right]_+,\\
        R_i
        &\leftarrow o^{-1}\!\left(
        \pi_i^{\mathrm{hist}}+\chi_i
        \left[o(S_i^{\mathrm{final}})-\pi_i^{\mathrm{hist}}\right]_+
        \right).
    \end{aligned}
    \label{eq:reliability-renewal}
\end{equation}
Here, $P_i^{\mathrm{init}}$, $P_i^{\mathrm{final}}$, and
$S_i^{\mathrm{final}}$ denote the initial pressure, final pressure, and final
support, respectively. $o$ and $o^{-1}$ denote the odds transformation and
its inverse, respectively, with $\pi_i^{\mathrm{hist}}=o(R_i)$.
The historical weight moves toward the precision
represented by final support, with positive growth gated by $\chi_i$.
Reliability increases only when adaptation reduces pressure and establishes
support exceeding its previous level. Each Gaussian learner enters the next
state with its adapted parameters and updated reliability.

\section{Experiments}
\label{sec:experiments}

\subsection{Experimental Setup}

\begin{table*}[t]
    \caption{\textbf{Novel-view synthesis results on CL-3D.} For each scene, results are averaged equally across evaluated states. Higher is better for PSNR and SSIM. Best results are bold, and second-best results are underlined.}
    \label{tab:clbench-main}
    \centering
    \fontsize{8pt}{10pt}\selectfont
    \setlength{\tabcolsep}{1.5pt}
    \begin{tabular*}{\textwidth}{@{\extracolsep{\fill}}l*{12}{c}@{}}
        \toprule
        & \multicolumn{2}{c}{ClubRoom}
          & \multicolumn{2}{c}{FloorLounge}
          & \multicolumn{2}{c}{Fridge}
          & \multicolumn{2}{c}{OfficeHub}
          & \multicolumn{2}{c}{StudyNook}
          & \multicolumn{2}{c}{ApartmentRoom} \\
        \cmidrule(lr){2-3}\cmidrule(lr){4-5}\cmidrule(lr){6-7}
        \cmidrule(lr){8-9}\cmidrule(lr){10-11}\cmidrule(lr){12-13}
        Method
          & PSNR $\uparrow$ & SSIM $\uparrow$
          & PSNR $\uparrow$ & SSIM $\uparrow$
          & PSNR $\uparrow$ & SSIM $\uparrow$
          & PSNR $\uparrow$ & SSIM $\uparrow$
          & PSNR $\uparrow$ & SSIM $\uparrow$
          & PSNR $\uparrow$ & SSIM $\uparrow$ \\
        \midrule
        3DGS~\cite{kerbl2023gaussians}~\textcolor{gray}{\scriptsize (TOG '23)} & 19.00 & 0.732 & 20.02 & 0.837 & 20.16 & 0.762 & 25.28 & 0.831 & 9.75 & 0.605 & 16.48 & 0.655 \\
        CLNeRF~\cite{cai2023clnerf}~\textcolor{gray}{\scriptsize (ICCV '23)}
        & \underline{22.33} & \underline{0.833} & 25.51 & 0.920 & \underline{24.95} & 0.831 & \underline{26.23} & \underline{0.881} & 21.49 & 0.809 & \underline{20.87} & \underline{0.769} \\
        3DGS + EWC~\cite{kirkpatrick2017ewc} & 19.88 & 0.793 & 22.90 & 0.933 & 24.61 & \underline{0.856} & 24.96 & 0.856 & 20.64 & 0.863 & 17.21 & 0.762 \\
        3DGS + MAS~\cite{aljundi2018mas} & 20.13 & 0.796 & 24.25 & 0.940 & 24.51 & \underline{0.856} & 25.08 & 0.867 & 20.67 & 0.863 & 17.15 & 0.764 \\
        CL-Splats~\cite{ackermann2025clsplats}~\textcolor{gray}{\scriptsize (ICCV '25)}
        & 20.40 & 0.827 & \underline{28.14} & \underline{0.941} & 21.91 & 0.773 & 20.62 & 0.818 & \underline{23.39} & \underline{0.877} & 20.37 & 0.760 \\
        O-SCD~\cite{galappaththige2026oscd}~\textcolor{gray}{\scriptsize (CVPR '26)}
        & 20.78 & 0.820 & 21.89 & 0.866 & 22.43 & 0.825 & 25.45 & 0.864 & 13.61 & 0.731 & 17.71 & 0.758 \\
        \midrule
        \textbf{\method} & \textbf{26.02} & \textbf{0.907} & \textbf{31.10} & \textbf{0.961} & \textbf{26.39} & \textbf{0.878} & \textbf{27.94} & \textbf{0.921} & \textbf{25.00} & \textbf{0.898} & \textbf{25.30} & \textbf{0.860} \\
        \bottomrule
    \end{tabular*}
\end{table*}

\paragraph{Datasets and Metrics.}
We evaluate our method on three real-world datasets for continual scene reconstruction.
The CL-Splats dataset~\cite{ackermann2025clsplats} focuses on local updates from sparse observations, with an initial state and one update state per scene.
The World Across Time (WAT) dataset~\cite{cai2023clnerf} provides dense observations across multiple states.
We conduct experiments on four scenes from each of these two datasets. 

Existing datasets, however, offer limited coverage of mixed geometric and appearance changes under incomplete revisits.
Their evaluation also primarily focuses on updated regions and provides a less direct assessment of how well methods preserve historical content beyond current observation coverage.
To better assess adaptation and preservation under mixed scene changes, we introduce the Continual-Learning-3D (CL-3D) dataset, comprising six real-world scenes.
They cover scene expansion, object additions, removals and rearrangements, and illumination variations, combining geometric and appearance changes.
Further dataset details are provided in Appendix~C.

We report PSNR and SSIM~\cite{wang2004ssim} to evaluate novel-view synthesis (NVS) quality.
For the CL-3D dataset, we explicitly divide the test views into changed and unchanged subsets to separately evaluate adaptation to revised scene content and preservation of still-valid content.

\paragraph{Baselines.}
We compare \method against representative 3D continual learning baselines.
\begin{itemize}
    \item \textbf{Selective local optimization.}
    Both detect changes in 2D to guide local updates.
    CL-Splats~\cite{ackermann2025clsplats} backprojects change masks, while
    O-SCD~\cite{galappaththige2026oscd} uses multi-view fusion before selective reconstruction and global refinement.

    \item \textbf{Replay.}
    CLNeRF~\cite{cai2023clnerf} preserves previous knowledge through replayed historical supervision.

    \item \textbf{Parameter regularization.}
    3DGS+EWC~\cite{kirkpatrick2017ewc} and 3DGS+MAS~\cite{aljundi2018mas} are constructed by augmenting the training objective with another regularization term.
    3DGS+EWC adds a quadratic penalty weighted by diagonal Fisher information, while 3DGS+MAS weights the penalty using parameter importance estimated from the sensitivity of rendered outputs.
\end{itemize}
We also report 3DGS~\cite{kerbl2023gaussians}, which directly fine-tunes the existing scene using only current observations.
All Gaussian-based methods share the same initial reconstruction and evaluation protocols.

\paragraph{Implementation Details.}
The initial reconstruction is optimized for 30,000 iterations.
Each subsequent state uses 10 rounds of 300 iterations on CL-Splats or 1,000 on CL-3D and WAT.
Support, pressure, and learning weights are recomputed from all current-state training views before the first round and after each round, remaining fixed during the following round.
We retain opacity pruning and densification for adaptive control of the Gaussian population~\cite{kerbl2023gaussians}, where pruning removes the reliability and splitting copies the parent's reliability to each child.
More details are provided in Appendix~B.
All experiments use a single NVIDIA GeForce RTX 4090 GPU.

\begin{figure}[t]
    \centering
    \includegraphics[width=\columnwidth]{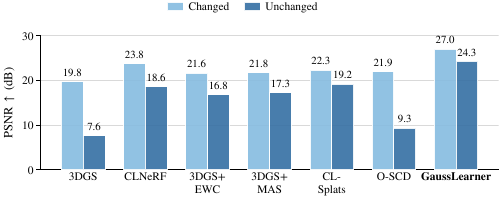}
    \caption{\textbf{Comparison of adaptation and preservation.}
    We report PSNR on changed and unchanged test views across all six scenes in the CL-3D dataset.
    Changed views measure adaptation to current observations, while unchanged views measure preservation of still-valid historical content.}
    \label{fig:changed-unchanged}
\end{figure}

\subsection{Main Results}

The main comparisons evaluate the central shift introduced by \method: viewing
3D Gaussian primitives as continual learners, rather than merely as targets of
an update policy.
Across 14 scenes from the CL-3D, CL-Splats, and WAT datasets, \method
improves average PSNR by 3.41 dB over the strongest baseline, O-SCD.
On CL-3D, \method consistently achieves the best
reconstruction quality (Table~\ref{tab:clbench-main}) and leads on both changed and unchanged test views,
demonstrating a strong balance between adaptation and preservation
(Fig.~\ref{fig:changed-unchanged}).
Qualitative comparisons in Fig.~\ref{fig:rivalsplats-qualitative} further
illustrate these improvements.

\begin{figure*}[t]
    \centering
    \includegraphics[width=\textwidth]{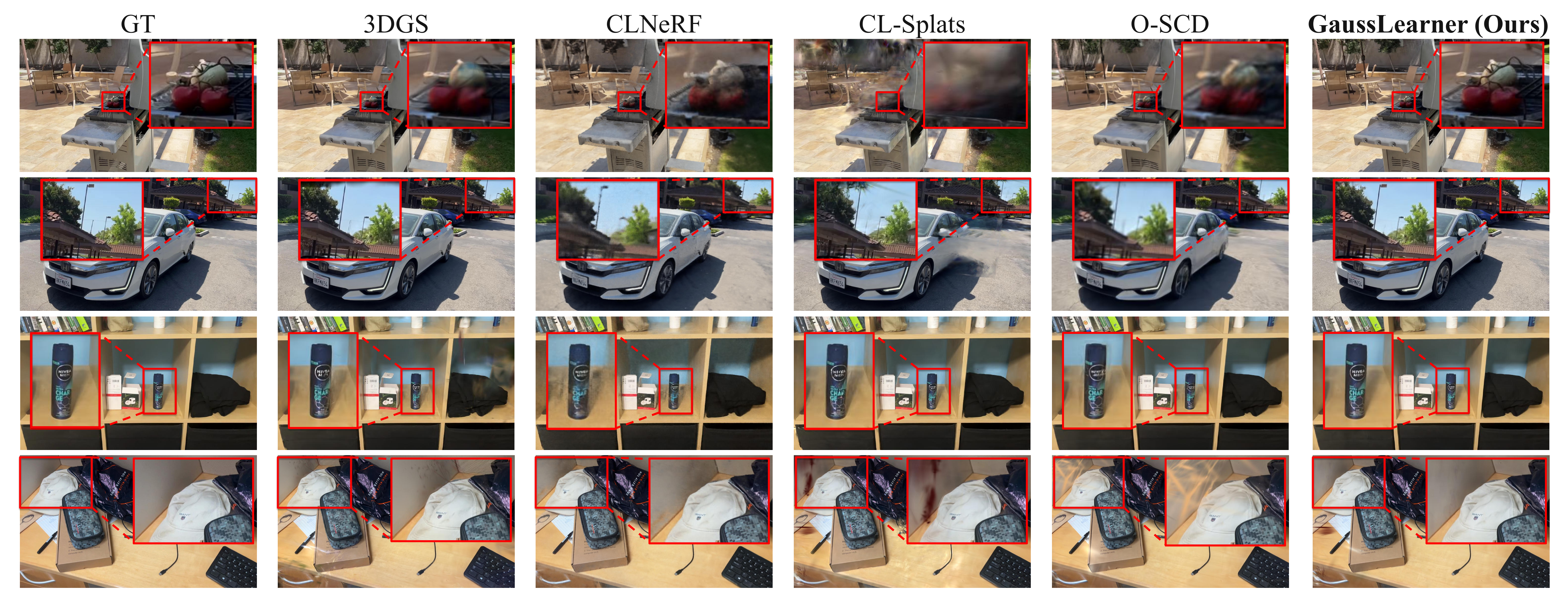}
    \caption{\textbf{Qualitative comparisons on WAT and CL-Splats.}
    From top to bottom are Grill, Car, Cupboard, and Desk.
    Red boxes and enlarged crops highlight how \method incorporates revised
    scene content while preserving surrounding content that remains valid.}
    \label{fig:qualitative-detailed}
\end{figure*}

\begin{table}[t]
    \caption{\textbf{Novel-view synthesis results on CL-Splats.} State 1 results are reported. Best results are bold, and second-best results are underlined.}
    \label{tab:clsplats-results}
    \centering
    \scriptsize
    \setlength{\tabcolsep}{1pt}
    \resizebox{\columnwidth}{!}{%
    \begin{tabular}{@{}l*{8}{c}@{}}
        \toprule
        & \multicolumn{2}{c}{Cupboard} & \multicolumn{2}{c}{Desk} & \multicolumn{2}{c}{Room} & \multicolumn{2}{c}{Shelf} \\
        \cmidrule(lr){2-3}\cmidrule(lr){4-5}\cmidrule(lr){6-7}\cmidrule(l){8-9}
        Method
        & PSNR $\uparrow$ & SSIM $\uparrow$
        & PSNR $\uparrow$ & SSIM $\uparrow$
        & PSNR $\uparrow$ & SSIM $\uparrow$
        & PSNR $\uparrow$ & SSIM $\uparrow$ \\
        \midrule
        3DGS~\cite{kerbl2023gaussians} & 25.69 & 0.918 & 28.32 & 0.875 & 20.93 & 0.725 & 26.52 & 0.914 \\
        CLNeRF~\cite{cai2023clnerf} & 30.21 & 0.916 & 29.78 & 0.860 & 29.92 & 0.937 & 22.48 & 0.867 \\
        3DGS + EWC~\cite{kirkpatrick2017ewc} & 30.59 & 0.954 & 27.68 & 0.884 & 21.05 & 0.731 & 29.64 & 0.939 \\
        3DGS + MAS~\cite{aljundi2018mas} & 30.89 & 0.956 & 27.51 & 0.883 & 21.09 & 0.732 & 29.24 & \underline{0.940} \\
        CL-Splats~\cite{ackermann2025clsplats} & 29.13 & 0.940 & 24.17 & 0.831 & 26.82 & 0.909 & 26.78 & 0.906 \\
        O-SCD~\cite{galappaththige2026oscd} & \underline{31.05} & \underline{0.957} & \underline{30.24} & \underline{0.900} & \underline{32.88} & \underline{0.966} & \underline{29.70} & \textbf{0.943} \\
        \midrule
        \textbf{\method} & \textbf{32.62} & \textbf{0.963} & \textbf{31.63} & \textbf{0.903} & \textbf{33.43} & \textbf{0.967} & \textbf{30.05} & \textbf{0.943} \\
        \bottomrule
    \end{tabular}}
\end{table}

\begin{table}[t]
    \caption{\textbf{Novel-view synthesis results on WAT.} For each scene, results are averaged equally across states 1--4. Best results are bold, and second-best results are underlined.}
    \label{tab:wat-results}
    \centering
    \scriptsize
    \setlength{\tabcolsep}{1pt}
    \resizebox{\columnwidth}{!}{%
    \begin{tabular}{@{}l*{8}{c}@{}}
        \toprule
        & \multicolumn{2}{c}{Breville} & \multicolumn{2}{c}{Car} & \multicolumn{2}{c}{Grill} & \multicolumn{2}{c}{Mac} \\
        \cmidrule(lr){2-3}\cmidrule(lr){4-5}\cmidrule(lr){6-7}\cmidrule(l){8-9}
        Method
        & PSNR $\uparrow$ & SSIM $\uparrow$
        & PSNR $\uparrow$ & SSIM $\uparrow$
        & PSNR $\uparrow$ & SSIM $\uparrow$
        & PSNR $\uparrow$ & SSIM $\uparrow$ \\
        \midrule
        3DGS~\cite{kerbl2023gaussians} & 30.60 & \textbf{0.942} & \underline{27.80} & \textbf{0.871} & \textbf{27.46} & \underline{0.866} & 31.02 & 0.949 \\
        CLNeRF~\cite{cai2023clnerf} & 29.20 & 0.870 & 22.54 & 0.590 & 24.13 & 0.675 & 28.46 & 0.917 \\
        3DGS + EWC~\cite{kirkpatrick2017ewc} & 30.35 & \textbf{0.942} & 27.27 & 0.851 & \underline{27.28} & \textbf{0.869} & \underline{31.33} & \textbf{0.952} \\
        3DGS + MAS~\cite{aljundi2018mas} & 30.19 & 0.939 & 27.19 & 0.847 & 26.96 & 0.864 & 31.30 & \underline{0.951} \\
        CL-Splats~\cite{ackermann2025clsplats} & 22.52 & 0.837 & 22.70 & 0.769 & 19.08 & 0.721 & 25.87 & 0.904 \\
        O-SCD~\cite{galappaththige2026oscd} & \underline{30.70} & 0.937 & 27.29 & \underline{0.866} & 26.13 & 0.845 & 30.28 & 0.946 \\
        \midrule
        \textbf{\method} & \textbf{31.57} & \underline{0.940} & \textbf{28.33} & \textbf{0.871} & 27.14 & 0.862 & \textbf{31.38} & \textbf{0.952} \\
        \bottomrule
    \end{tabular}}
\end{table}

On CL-Splats, \method achieves the highest PSNR across all four scenes
(Table~\ref{tab:clsplats-results}), showing
robust adaptation from sparse local observations.
On WAT, it remains competitive across repeated state changes
(Table~\ref{tab:wat-results}).
Grill provides dense and comprehensive observations, under which full-scene
optimization of 3DGS already reconstructs new states effectively, leaving less
room for continual-learning mechanisms.
Together, these results show that \method remains effective across both sparse
local observations and repeated multi-state changes.
Detailed comparisons are shown in Fig.~\ref{fig:qualitative-detailed}.

The results on the three datasets, together with the changed and unchanged view splits in CL-3D,
reveal distinct behaviors among the baselines.
Direct 3DGS~\cite{kerbl2023gaussians} fine-tuning sharply degrades on unchanged
views, exposing catastrophic forgetting due to overfitting to current observations.
Replay with CLNeRF~\cite{cai2023clnerf} improves both adaptation and
preservation, whereas parameter regularization partially protects unchanged
content at the cost of restricting necessary changes.
CL-Splats~\cite{ackermann2025clsplats} achieves the strongest unchanged-view
quality among the baselines, consistent with its selective local optimization, but
performs less favorably on changed views.
This limitation is also reflected in its reconstruction under the complex
occlusions in OfficeHub and appearance variations in Fridge, where reliable 2D
change detection becomes difficult.
O-SCD~\cite{galappaththige2026oscd} similarly isolates local updates, but its
final global optimization exposes unchanged content to current observations and
weakens preservation.
More quantitative results, efficiency analysis, and additional qualitative
comparisons are provided in Appendices~E.1, E.3, and H, respectively.

\method continuously regulates how much each Gaussian changes using current pressure and cross-state reliability.
This allows adaptation without restricting updates to detected change masks, while protecting still-valid content from excessive adaptation.
On CL-3D, geometric changes, illumination changes, and incomplete observations occur together, highlighting the advantage of jointly regulating adaptation and preservation at the Gaussian level.

\subsection{Learning Dynamics Analysis}

\begin{figure}[t]
    \centering
    \includegraphics[width=\columnwidth]{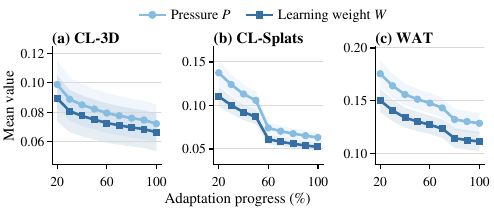}
    \caption{\textbf{Within-state self-regulation.}
    Mean pressure $P$ and learning weight $W$ over normalized update progress.
    Shading denotes standard error.}
    \label{fig:within-state-self-regulation}
\end{figure}

Across all three datasets, mean pressure $P$ and learning weight $W$ decrease
as adaptation proceeds (Fig.~\ref{fig:within-state-self-regulation}), indicating
that current observations become better explained while parameter adaptation
and optimizer-memory refresh weaken.
These trends are consistent with within-state self-regulation that reduces
adaptation strength as observations are incorporated.
Appendix~E.2 further illustrates reliability renewal at scene-state
boundaries, connecting adaptation outcomes to learning in subsequent scene
states.

\subsection{Ablations}

We examine observation aggregation, learning-weight construction, parameter
adaptation, optimizer-memory refresh, and reliability renewal within
\method's self-regulating optimization process
(Tables~\ref{tab:component-ablation} and~\ref{tab:optimizer-memory-ablation}).

\begin{table}[t]
    \caption{\textbf{Ablating Observation Aggregation, Self-Regulation, Parameter Adaptation, and Reliability Renewal.}
    Results are equally averaged over StudyNook, Fridge, and ClubRoom.}
    \label{tab:component-ablation}
    \centering
    \scriptsize
    \setlength{\tabcolsep}{3.0pt}
    \begin{tabular*}{\columnwidth}{@{\extracolsep{\fill}}lcccc@{}}
        \toprule
        & \multicolumn{2}{c}{Changed}
          & \multicolumn{2}{c}{Unchanged} \\
        \cmidrule(lr){2-3}\cmidrule(l){4-5}
        Variant
        & PSNR $\uparrow$ & SSIM $\uparrow$
        & PSNR $\uparrow$ & SSIM $\uparrow$ \\
        \midrule
        View-Wise Sum
        & 26.10 & 0.901
        & 21.86 & 0.857 \\
        $W=P$
        & 26.02 & 0.901
        & 21.90 & 0.859 \\
        Pre-Adam Gradient Scaling
        & 25.90 & 0.897
        & 21.75 & 0.851 \\
        Frozen Reliability
        & 26.06 & 0.902
        & 21.87 & 0.860 \\
        Ungated Reliability Renewal
        & 26.05 & 0.902
        & 21.88 & 0.860 \\
        \midrule
        \textbf{\method}
        & \textbf{26.44} & \textbf{0.903}
        & \textbf{22.18} & \textbf{0.861} \\
        \bottomrule
    \end{tabular*}
\end{table}

\paragraph{Direction-Aware Pooling.}
View-Wise Sum replaces our directional pooling operator with a sum of per-view
support and pressure followed by the same normalization, which lowers
performance on both changed and unchanged views.
Directionally balanced aggregation better captures geometric diversity across viewing directions,
thus providing a stronger observational basis for Gaussian self-regulation.

\paragraph{Reliability-Aware Self-Regulation.}
Setting $W=P$ removes the reliability-dependent adaptation share $\beta$,
making attributed pressure alone determine the learning weight for both
parameter adaptation and optimizer-memory refresh.
Its lower performance on both view splits shows that cross-state reliability
helps regulate how strongly an inconsistency in current observations should drive learning.

\paragraph{Parameter Adaptation.}
Pre-Adam Gradient Scaling applies $W$ to gradients before Adam, where moment
normalization can alter its intended adaptation scale ~\cite{kingma2015adam}. Its lower performance
supports acting on the completed Adam proposal,
where $W$ directly controls how much parameter change each Gaussian realizes.

\begin{table}[t]
    \caption{\textbf{Ablating Optimizer Memory Refresh for Subsequent Adaptation.}
    Results are equally averaged over StudyNook, Fridge, and ClubRoom.
    $\bar\gamma$ is the mean assigned boundary-retention factor;
    $\rho_m$ and $\rho_{\sqrt{v}}$ are the geometric-mean RMS retention of the
    first- and square-root second-moment states.}
    \label{tab:optimizer-memory-ablation}
    \centering
    \scriptsize
    \setlength{\tabcolsep}{2.0pt}
    \begin{tabular*}{\columnwidth}{@{\extracolsep{\fill}}lccccc@{}}
        \toprule
        Variant & Changed PSNR $\uparrow$ & Unchanged PSNR $\uparrow$
        & $\bar\gamma$ & $\rho_m$ & $\rho_{\sqrt{v}}$ \\
        \midrule
        Dense Adam
        & 25.71 & 21.95 & \multicolumn{3}{c}{No boundary refresh} \\
        Sparse Adam~\cite{mallick2024taming}
        & 25.71 & 21.37 & \multicolumn{3}{c}{No boundary refresh} \\
        \midrule
        \shortstack[l]{Uniform Refresh\\ $\gamma=0.90$}
        & 25.68 & 21.98 & 0.900 & 0.900 & 0.900 \\
        $\gamma=0.95$
        & 25.67 & 21.97 & 0.950 & 0.950 & 0.950 \\
        \midrule
        \textbf{\method}
        & \textbf{26.44} & \textbf{22.18} & 0.944 & 0.816 & 0.829 \\
        \bottomrule
    \end{tabular*}
\end{table}

\paragraph{Optimizer Memory Refresh.}
Table~\ref{tab:optimizer-memory-ablation} compares optimizer-memory refresh policies.
Dense Adam and Sparse Adam~\cite{mallick2024taming} retain unattenuated moment updates even when a
Gaussian accepts only part of the parameter proposal, while Uniform Refresh
applies the same retention to every Gaussian. Weight-Guided Optimizer Memory
Refresh instead aligns retention with each Gaussian's learning weight and
achieves the highest PSNR. The resulting moment RMS retention is lower than the mean assigned
retention, suggesting that our adaptive policy applies stronger attenuation to
larger accumulated moments under high pressure, limiting their influence on
subsequent adaptation.

\paragraph{Cross-State Reliability Renewal.}
Frozen Reliability keeps $R$ fixed across states, whereas Ungated Reliability
Renewal updates it from stronger final support regardless of whether current
pressure is resolved. Both yield lower reconstruction quality, supporting
the renewal rule that combines pressure resolution with stronger final support.

\section{Limitations}
\label{sec:limitations}

Despite its effectiveness in balancing adaptation and preservation, \method has several limitations.
It remains sensitive to challenging illumination conditions, including specular reflections and refraction, as illustrated in Appendix~E.4.
Incorporating an inverse rendering framework may improve robustness.
Moreover, \method maintains a reconstruction of the latest scene state.
The representation does not support arbitrary replay of previous states.
Object-centric state management could extend \method toward historical-state retrieval.

\section{Conclusion}
\label{sec:conclusion}

We present \method, a continual learning framework that views 3D Gaussian primitives as self-regulating continual learners rather than merely as update targets.
Each Gaussian balances current observations against cross-state reliability through a learning weight that governs both parameter adaptation and optimizer-memory refresh.
Adaptation outcomes then renew its reliability to guide future learning.
We also introduce CL-3D to evaluate continual reconstruction under mixed scene changes and incomplete observations, with separate assessment of adaptation and preservation.
Across 14 scenes, \method improves average PSNR by 3.41 dB over the strongest baseline, O-SCD.
Together with its balance between adaptation and preservation, these results demonstrate the benefits of this continual-learning paradigm.
Ablations further validate the effectiveness of our designs within the closed continual-learning loop.

\clearpage
{
    \small
    \bibliographystyle{ieeenat_fullname}
    \bibliography{main}
}


\end{document}